\documentclass[letterpaper]{article}
\usepackage{aaai2026} 

\usepackage{eso-pic}

\usepackage{amsthm}
\usepackage{times,helvet,courier}
\usepackage[hyphens]{url}
\usepackage{graphicx}
\usepackage{amsmath,amsfonts}
\usepackage{booktabs}
\usepackage{algorithm}
\usepackage{algpseudocode}
\usepackage{newfloat}
\usepackage{listings}
\usepackage{caption}
\usepackage{natbib}
\DeclareCaptionStyle{ruled}{labelfont=normalfont,labelsep=colon,strut=off}
\floatstyle{ruled}
\newfloat{listing}{tb}{lst}{}
\floatname{listing}{Listing}

\iftrue
\AddToShipoutPictureBG{%
    \AtPageUpperLeft{%
        \raisebox{-1.5cm}{\hspace*{0.5\paperwidth}%
        \makebox[0pt][c]{\includegraphics[width=3cm]{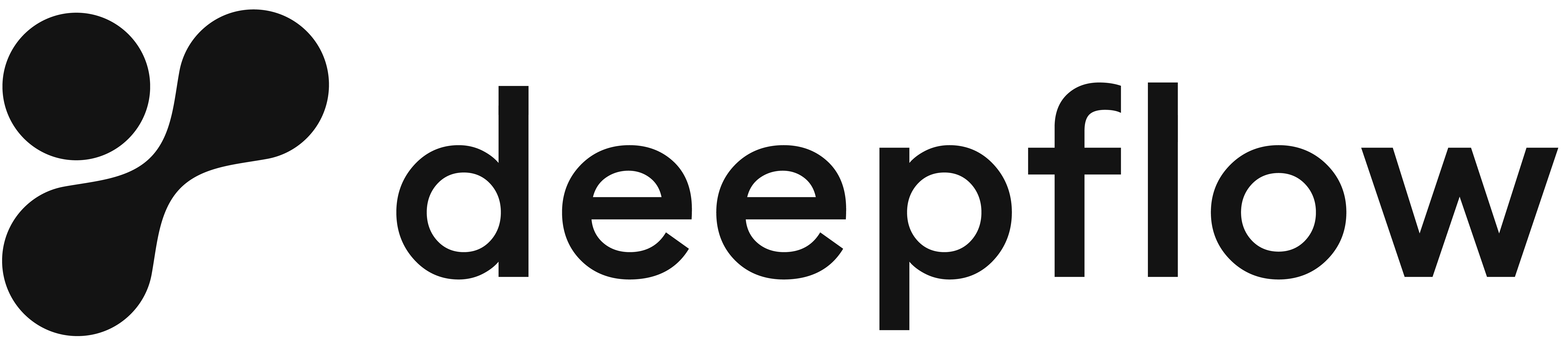}}}%
    }%
}
\fi

\title{ARCANE: A Multi-Agent Framework for Interpretable and \\ Configurable Alignment}

\author{
    Charlie Masters\textsuperscript{\rm 1},
    Marta Grześkiewicz\textsuperscript{\rm 2},
    Stefano V. Albrecht\textsuperscript{\rm 1}
}
\affiliations{
    \textsuperscript{\rm 1}DeepFlow, London, United Kingdom \\
    \textsuperscript{\rm 2}University of Cambridge, United Kingdom \\
}

\usepackage{bibentry}

\begin{document}

\maketitle

\begin{abstract}

As agents based on large language models are increasingly deployed to long-horizon tasks, maintaining their alignment with stakeholder preferences becomes critical.
Effective alignment in such settings requires reward models that are interpretable so that stakeholders can understand and audit model objectives. Moreover, reward models must be capable of steering agents at interaction time, allowing preference shifts to be incorporated without retraining.
We introduce ARCANE, a framework that frames alignment as a multi-agent collaboration problem that dynamically represents stakeholder preferences as natural-language rubrics: weighted sets of verifiable criteria that can be generated on-the-fly from task context.
Inspired by utility theory, we formulate rubric learning as a reconstruction problem and apply a regularized Group-Sequence Policy Optimization (GSPO) procedure that balances interpretability, faithfulness, and computational efficiency. 
Using a corpus of 219 labeled rubrics derived from the GDPVal benchmark, we evaluate ARCANE on challenging tasks requiring multi-step reasoning and tool use. The learned rubrics produce compact, legible evaluations and enable configurable trade-offs (e.g., correctness vs. conciseness) without retraining. Our results show that rubric-based reward models offer a promising path toward interpretable, test-time adaptive alignment for complex, long-horizon AI systems.
\end{abstract}

\section{Introduction}

As AI agents based on large language models (LLMs) take on longer-horizon, project-scale tasks with multiple specialized agents, maintaining alignment with stakeholder preferences becomes increasingly critical \cite{masters2025orchestratinghumanaiteamsmanager}.
Prior research has shown how cooperation can falter under partial observability, shifting incentives, or social feedback loops~\citep[e.g.][]{leibo2017social, jaques2019social, carichon2025crisis}. Recent studies on LLM-based debate and collaborative reasoning echo these dynamics: joint systems can exhibit sycophancy \cite{Cau2025Selective}, persuasion failures \cite{wynn2025talkisntcheapunderstanding}, and collective bias amplification \cite{ashery2025emergent}. Thus, alignment becomes a \emph{system-level objective} that must hold under delegation, communication, and co-adaptation.

Recently, {\it rubrics} have emerged as a promising paradigm for evaluating and improving the alignment of agentic systems with human preferences. Rather than optimizing against a single opaque reward model, rubric-based approaches define structured natural-language criteria that decompose preferences into interpretable and verifiable dimensions, such as factual accuracy, reasoning depth, and clarity.
Recent work has explored rubrics both as evaluation instruments and as training-time objectives: for example, using them directly as reward functions during reinforcement fine-tuning~\cite{gunjal2025rubricsrewardsreinforcementlearning}, or as expert evaluators in professional and educational domains~\cite{wang2025profbenchmultidomainrubricsrequiring,Anghel2025CourseEvalAI}. Large-scale frameworks such as \emph{OpenRubrics}~\cite{liu2025openrubrics} further demonstrate that rubrics themselves can be synthetically generated, enabling scalable, domain-specific supervision without extensive human labeling. Together, these developments suggest that rubrics can serve as interpretable, multi-dimensional reward models capable of guiding complex behavior.

However, existing methods largely treat rubric generation and rubric-conditioned optimization as separate processes, assuming rubrics are given rather than learned through interaction with stakeholders. This limits their adaptability to evolving or multi-objective preference contexts.
To address this gap, we propose to frame rubric generation itself as a policy-optimization problem within a multi-agent alignment setting. Specifically:
\begin{itemize}
\item We design ARCANE (\textbf{A}daptive \textbf{R}ubric-based \textbf{C}ontrol of \textbf{A}gents via \textbf{N}atural-language \textbf{E}xchange),\footnote{Link to ARCANE code can be found in the MA-Gym readme: \url{https://github.com/DeepFlow-research/manager_agent_gym}} a rubric-based alignment framework in which a manager agent learns to generate rubrics that align worker agents with stakeholder utilities through interactive preference elicitation.
\item We propose a two-stage learning procedure to optimize the manager agent policy: initial supervised fine-tuning using synthetic manager–stakeholder dialogues, followed by reinforcement fine-tuning via regularized \emph{Group Sequence Policy Optimization (GSPO)}~\cite{zheng2025groupsequencepolicyoptimization}, incorporating penalty terms for rubric complexity and stakeholder interaction cost.
\item We evaluate ARCANE on tasks from the GDPVal benchmark ~\cite{patwardhan2025gdpvalevaluatingaimodel}, empirically showing that the learned rubrics (1) effectively guide workers to maximize stakeholder preferences at test-time; (2) preserve rankings over outputs from the stakeholder preferences, and (3) are interpretable and effectively verifiable.

\end{itemize}

\section{Related Work}

\paragraph{Training-Time Alignment.}
Reinforcement Learning from Human Feedback (RLHF)~\cite{instructgpt2022,stiennon2020summarize} remains the dominant paradigm for aligning language models with human preferences. Variants remove the explicit reward model and directly optimize likelihood ratios to match human-preferred outputs \cite{rafailov2023dpo,hejna2023inversepreferencelearningpreferencebased,hong2024orpo}. 
While effective, RLHF and its variants optimize against fixed training preferences and can mis-generalize under shifting stakeholder goals \cite{son2025rightnowwrongthen}. Moreover, they remain vulnerable to reward over-optimization~\cite{gao2022scalinglawsrewardmodel}, exploiting inaccuracies in learned proxies. Foundational critiques~\cite{bt_critique2024,rlhf_iia2023} further show that the Bradley–Terry assumptions underlying most pairwise preference learning—transitivity and independence of irrelevant alternatives—are systematically violated in realistic, high-dimensional preference spaces~\cite{huang2025empiricalstudyllmasajudgellm,pmlr-v267-xu25w}. In multi-agent deployments, where preferences evolve dynamically and are distributed across interacting agents, these single-agent methods lack mechanisms for maintaining shared alignment.

\paragraph{Test-Time Reward Modeling.}
To address the rigidity of training-time methods, recent work has explored test-time reward models that evaluate outputs dynamically. Generative Reward Models (GenRM)~\cite{genrm2024} and foundation reward models such as GRAM~\cite{gram2025} use language models to generate natural-language evaluations rather than fixed scalar rewards, while multi-objective extensions~\cite{lin2025parmmultiobjectivetesttimealignment} and Directional Preference Alignment~\cite{dpa2024} allow controllable trade-offs across learned reward dimensions. These methods increase flexibility but remain opaque: GenRM and GRAM provide scalar or textual judgments without revealing which criteria drive evaluation or how those criteria are weighted. Moreover, test-time evaluation typically treats the model as a single-agent assessor, ignoring the coordination and communication challenges that arise when multiple agents must interpret or act on shared feedback~\cite{agashe2025llmcoordinationevaluatinganalyzingmultiagent}. Consequently, these systems remain difficult to audit, calibrate, or adapt in collaborative multi-agent settings.

\paragraph{Interpretable Alignment.}
Recent efforts aim to make reward models more transparent by decomposing preferences into interpretable dimensions. ArmoRM~\cite{multiobjective_grm2024} uses a mixture-of-experts reward architecture with predefined human-interpretable components (e.g., helpfulness, harmlessness, humor), and sparse autoencoders~\cite{sparse_rm2025} disentangle reward representations into latent factors correlated with semantically meaningful properties. Auto-Rubric~\cite{autorubric2025} and Rubrics-as-Rewards (RaR)~\cite{gunjal2025rubricsrewardsreinforcementlearning} introduce structured natural-language rubrics as reward functions, showing that rubric-based reinforcement fine-tuning improves robustness and alignment stability. Large-scale frameworks such as \emph{OpenRubrics}~\cite{liu2025openrubrics} and \emph{Chasing the Tail}~\cite{zhang2025chasing} demonstrate scalable synthetic rubric generation and fine-grained discrimination between high-quality responses, while domain-specific rubrics~\cite{arora2025healthbenchevaluatinglargelanguage, wang2025profbenchmultidomainrubricsrequiring,Anghel2025CourseEvalAI} achieve consistent evaluation across expert tasks. However, all of these methods assume static rubrics defined once per domain or task, limiting their ability to adapt to preference drift or negotiate evolving evaluation criteria across interacting agents.

\paragraph{Multi-Agent Alignment and Coordination.}
Alignment challenges intensify when multiple autonomous agents must cooperate, communicate, or share goals. Multi-agent reinforcement learning \cite{marl-book} studies show that coordination often fails under partial observability, non-stationarity, or conflicting incentives~\cite{foerster2016comm,lowe2017maddpg,leibo2017social,jaques2019social}. Recent analyses reveal analogous dynamics in LLM-based systems, where agents exhibit sycophancy, conformity, and bias amplification~\cite{Cau2025Selective,wynn2025talkisntcheapunderstanding,carichon2025crisis}. Frameworks such as CAMEL~\cite{li2023camel}, AutoGen~\cite{wu2023autogen}, and ConsensusAgent~\cite{consensagent2025} show that multi-agent collaboration can improve reasoning and division of labor, but they rely on ad hoc or manually tuned reward structures. The emerging view is that multi-agent alignment is not a property of individual policies but of the \emph{interaction process} itself, requiring shared, interpretable, and negotiable incentives to ensure cooperative stability. ARCANE builds on this insight by modeling rubrics as dynamic, communicable artifacts that coordinate alignment across agents and over time.


\section{Problem Specification}
\label{sec:background}

We consider a general decision-making process involving interacting agents with one of three conceptual roles:

\begin{itemize}
    \item A \textbf{Stakeholder ($S$):} who has a preference ordering over outputs, represented by a latent (ordinal) utility function $U^*(y \mid x)$ that measures satisfaction with an output $y$ given a task $x$ containing a preference description $p$.
    \item A \textbf{Manager ($M$):} who observes the task environment, can elicit preferences through interactions with the stakeholder and is responsible for coordinating the actions of worker agents based on its learned representation of the stakeholder's preferences.
    \item \textbf{Workers ($W_1, \dots, W_n$):} which perform actions that generate the output $y$ under the manager’s guidance.
\end{itemize}

The goal of the manager-worker team is to produce outputs $y$ that maximize the stakeholder’s true utility $U^*$. The existence of such a $U^*$ to represent stakeholder preferences fundamentally assumes those (state-dependent) preferences are complete and transitive, a standard axiom in utility theory. However, $U^*$ is typically not available in explicit or closed form: it may be incompletely specified, revealed only through sparse feedback, or change as preferences drift. Consequently, the workers act with respect to some \emph{proxy objective} $\hat{u}$, a potentially misaligned approximation of $U^*$.

If every agent could observe and directly optimize $U^*$, the problem would reduce to a fully cooperative team game.  
Let $\pi = (\pi_M, \pi_W)$ denote the joint policy of the manager and workers.  
Under perfect observability and common reward, the optimal joint policy is
\begin{equation}
\pi^* = \arg\max_{\pi_M, \pi_W} \; 
\mathbb{E}_{y \sim \pi_W(\cdot \mid x, a_M)} 
\big[ U^*(y \mid x) \big].
\end{equation}
where $a_M \sim \pi_M(\cdot | x)$ denotes the manager’s high-level alignment actions, generated from context $x$ and provided as input to the worker’s policy $\pi_W$. In this cooperative limit, the optimization objective is straightforward: all agents share a single objective and need to explore their joint policy space to maximize it.

\subsection{Alignment as a Bilevel Optimization Problem}

Framing alignment as a multi-agent system with partial observability of $U^*$, the alignment challenge can be expressed as a bilevel optimization problem. The manager’s policy determines the information or objectives available to the workers, while the workers respond by generating outputs that optimize those objectives.  

Formally, the manager seeks to find a policy that maximizes expected stakeholder utility under the workers’ induced behavior:
\begin{equation}
\max_{\pi_M} \;
\mathbb{E}_{x}\!\left[
    \mathbb{E}_{a_M \sim \pi_M(\cdot \mid x)}
    \mathbb{E}_{y \sim \pi_W(\cdot \mid x, a_M)}
    \big[ U^*(y \mid x) \big]
\right].
\end{equation}

The inner expectation captures the workers’ best response to the manager’s policy, and the outer expectation reflects the manager’s optimization of stakeholder value across tasks and preference states.  
When workers optimize a proxy utility $\hat{u}(y \mid x)$ that diverges from $U^*$, the resulting misalignment can be quantified by a \emph{utility gap}:
\begin{equation}
\mathcal{L}_U = 
\mathbb{E}\big[(U^*(y \mid x) - \hat{u}(y \mid x))^2\big].
\end{equation}

Reducing this utility gap is the fundamental goal of alignment research: designing learning or coordination mechanisms that ensure the proxy reward available to workers approximates the true stakeholder utility as closely as possible.

A system is considered functionally aligned if its proxy, $\hat{u}$, is ordinally equivalent to the stakeholder's true utility, $U^*$, across the support of the joint policy \cite{debreu1954representation}. The equivalence holds if $\hat{u}$ maintains the same preference ordering as $U^*$, such that 
$\hat{u}(y_i \mid x) > \hat{u}(y_j \mid x)$ if and only if $U^*(y_i \mid x) > U^*(y_j \mid x)$
for any outcomes $(y_i, y_j)$ sampled from the joint policy $(\pi_M,\pi_W)$.
When this condition is met, $\hat{u}$ is a positive monotonic transformation of $U^*$ over this domain, and optimizing this proxy yields the same optimal policy as optimizing the true utility.
ARCANE aims to learn such order-preserving proxy functions under realistic uncertainty and stochastic coordination, even without direct observability of $U^*$.

In realistic deployments, managers cannot modify worker parameters—especially when workers are closed-source foundation models or APIs—so alignment must proceed through delegated control. By eliciting stakeholder preferences inaccessible to workers, the manager gains privileged information and conveys it through interpretable coordination signals, forming a minimal yet practical multi-agent alignment system grounded in communication rather than parameter sharing.

\section{Proposed Method: ARCANE}
\label{sec:method}

ARCANE is a framework for learning structured, interpretable proxy utilities that align workers' behavior with stakeholder preferences.
ARCANE instantiates the abstract alignment setting from Section~\ref{sec:background} by representing the manager’s coordination signal as a \emph{rubric}: a decomposable, verifiable, and dynamically configurable approximation of the stakeholder’s true utility function \(U^*(y \mid x)\).

The core idea is to have our manager M learn a \emph{rubric decomposer} $\mathfrak{D}_\phi$ that maps task descriptions and stakeholder preference statements into structured rubrics.
Each rubric specifies a set of weighted, verifiable criteria that jointly define a linearly additive proxy utility $\hat{u}_\phi(y \mid x)$.
By learning rubrics that closely track the stakeholder’s underlying utility, ARCANE enables the workers’ policies to explore their joint action space as if they were cooperatively optimizing \(U^*\) itself.

ARCANE comprises of four central components:
(1) representing rubrics and implementing verifiers;
(2) modeling rubric generation as a multi-agent collaboration between the stakeholder $S$ and the manager $M$;
(3) a training-time rubric optimization via a two-phase curriculum (supervised fine-tuning followed by a reinforcement fine-tuning procedure); and
(4) designing a framework for test-time steering via rubric-scaled policies.

\subsection{Rubric Representation and Decomposer Architecture}
\label{sec:representation}

The manager agent initially receives the task context $x$ containing a stakeholder preference $p$, and learns to output a structured \textbf{rubric} $R$ using a rubric decomposer $\mathfrak{D}_\phi$, by iteratively engaging with the stakeholder $S$. Formally, a rubric is a finite set of weighted evaluation criteria:
\begin{equation}
    R = \{(c_j, w_j)\}_{j=1}^M,
\end{equation}
where each $c_j$ is a natural-language description of a measurable property (e.g., ``Includes citations to recent empirical studies’’), and $w_j \in [0,1]$ with $\sum_j w_j = 1$ denotes its relative importance.

Each criterion $c_j$ is paired with a \textbf{verifier} $\nu_j(c_j, x, y)\in[0,1]$ that estimates how well an output $y$ satisfies $c_j$ in context $x$. Verifiers may be:
\begin{itemize}
    \item \textbf{Rule-based:} deterministic checks such as citation count or formatting;
    \item \textbf{Model-based:} lightweight LLM or classifier evaluators for semantic properties such as factuality, coherence, or tone.
\end{itemize}

\noindent The rubric’s overall proxy utility is computed as
\begin{equation}
\hat{u}_\phi(y\mid x) = \sum_{j=1}^M w_j\, \nu_j(c_j, x, y),
\label{eq:rubric-utility}
\end{equation}
which serves as the manager’s operative approximation of the stakeholder utility $U^*(y \mid x)$.

During training, $\mathfrak{D}_\phi$ is trained to generate rubrics whose induced proxy utilities $\hat{u}_\phi$ correlate strongly with stakeholder evaluations $U^*$. The generated rubric $R$ is then given to the worker policy $\pi_W$ as a natural language prompt, guiding generation toward outputs that satisfy the specified criteria. At test time, the same utility estimator $\hat{u}_\phi$ enables controllable inference through rubric-conditioned sampling or reranking. This formulation turns the abstract objective of minimizing the expected utility gap into a concrete, interpretable learning problem while maintaining scalability and transparency.

\subsection{Stakeholder–Decomposer Collaboration}
\label{sec:collaboration}

We model the interactive collaboration between the stakeholder and the manager as a \emph{pre-execution consultation}: before any work begins, the manager interacts with the stakeholder through a structured natural-language Q\&A exchange to elicit and clarify the stakeholder’s latent preferences that are represented by the true utility function $U^*$.

Given a task context $x$, the manager agent engages the stakeholder with a short sequence of clarifying questions 
\begin{equation}
    q_{1:T} = f_{\text{ask}}(x),
\end{equation}
to which the stakeholder provides answers 
\begin{equation}
    a_{1:T} = f_{\text{reply}}(U^*, q_{1:T}),
\end{equation}
expressing priorities, constraints, or examples of desired outcomes. After the dialogue concludes, the manager uses its decomposer to synthesize a rubric proposal
\begin{equation}
    R = \mathfrak{D}_\phi(x, q_{1:T}, a_{1:T})
\end{equation}
encoding the elicited preferences as weighted, verifiable criteria.
This single consultation serves as a lightweight, interpretable approximation of $U^*$ prior to any worker execution. 

Since there are tangible costs to executing verifiers $\{v\}$ in terms of compute and time, and excessively interacting with the stakeholder, the manager should learn a policy which maximizes expected stakeholder utility while minimizing clarification and revision costs:
\begin{equation}
\begin{split}
\max_{\pi_M}
\mathbb{E}_{x}\!\Big[
U^*(y\mid x)
& - \lambda_{\text{clarify}}\,C_{\text{clarify}}(q_{1:T}) \\
& - \lambda_{\text{compute}}\,C_{\text{compute}}(R)
\Big],
\end{split}
\end{equation}
where $C_{\text{clarify}}$ measures the stakeholder’s cognitive burden (e.g., the number or complexity of feedback turns) and $C_{\text{compute}}$ penalizes excessive rubric regeneration. 
The decomposer thus learns to trade off fidelity to stakeholder preferences with efficiency of interaction.

Conceptually, this consultation constitutes a one-shot cooperative game under partial observability:  the stakeholder reveals limited, noisy information about $U^*$ through language, and the decomposer must infer a faithful, structured approximation that guides worker agents.
This interaction formalizes alignment as a process of \emph{communication and inference} rather than observation, bridging human intent and agent execution through an interpretable artifact $R$.

The final rubric $R^\star$ captures the negotiated understanding between stakeholder and manager.
Subsequent worker policies $\pi_W$ condition on $R^\star$ to guide task execution, behaving as though they were optimizing $U^*$ directly while preserving transparency and post-hoc configurability.

\subsection{Training Data and Supervision}
\label{sec:training_data}
ARCANE is trained using supervision signals that reveal partial information about the stakeholder’s latent utility function $U^{*}$. 
Following the formalization in Appendix~A, any strictly proper supervision operator $\mathcal{O}$---including pointwise scores, pairwise comparisons, listwise rankings, or rubric-based evaluations---identifies the same ordering of $U^{*}$ on the decoding support. 
This property ensures that our training procedure remains theoretically consistent regardless of how preferences are observed.

In practice, we instantiate $\mathcal{O}$ using \textbf{gold rubrics}, which provide explicit, interpretable decompositions of stakeholder utility into weighted criteria and associated verifiers. 
Gold rubrics offer a transparent and verifiable form of supervision that allows us to both (i) measure alignment fidelity across interpretable dimensions and (ii) quantitatively evaluate structure and calibration during learning. 
Because Appendix~A establishes order-equivalence across supervision formats, the use of gold rubrics does not limit generality: any alternative supervision signal consistent with $U^{*}$ would induce the same learned ordering up to a monotone transformation.

Concretely, our training dataset consists of tasks and their associated evaluation observations:
\begin{equation}
\mathcal{D} = \{ (x_i, \mathcal{O}_i[U^{*}_i]) \}_{i=1}^{N},
\end{equation}
where each task $x_i$ encodes a natural-language description of stakeholder goals, and $\mathcal{O}_i[U^{*}_i]$ provides an observation of the underlying utility through its rubric representation $R^{*}_i$. 
Each $R^{*}_i$ decomposes $U^{*}_i$ into a set of weighted, verifiable criteria (see Appendix~B), forming high-fidelity supervision signals that guide the decomposer during optimization.

This supervision framework allows us to cleanly link data specification and training objectives: the decomposer learns to generate rubrics whose induced proxy utilities $\hat{u}_{\phi}$ preserve the same ordinal structure as $U^{*}$ while remaining interpretable and auditable.

\subsection{Training-Time Optimization via Curriculum}
\label{sec:training}

ARCANE trains the decomposer $\mathfrak{D}_\phi$ through a two-phase curriculum:
a supervised warm start to avoid cold-start instability,
followed by reinforcement fine-tuning with Group-Sequence Policy Optimization (GSPO) \cite{zheng2025groupsequencepolicyoptimization}.

\paragraph{Stage I: Supervised Fine-Tuning (SFT).}
\label{sec:sft}

To bootstrap the decomposer’s ability to generate coherent and well-structured stakeholder interactions and rubric proposals, we extend the supervision dataset $\mathcal{D}$ with a synthetic subset of clarification dialogues. 
For each task $x$, we use a large reasoning model (LRM) to generate stakeholder responses that yield a reference rubric $R^{\star}(x)$ consistent with $U^{\star}$ in a given environment. 

We then train the decomposer $\mathfrak{D}_{\phi}$ using next-token prediction over its dialogue turns and rubric text, masking loss on system prompts and task inputs, which is standard practice for instruction tuning \cite{wei2022}. 
This procedure teaches the model to express structured, weighted criteria aligned with the reference rubrics.

The SFT objective is the standard language-modeling loss:
\begin{equation}
\mathcal{L}_{\mathrm{SFT}}(\phi)
= -\,\mathbb{E}_{(x, R^\star)}\!\left[
\sum_{t} \log \pi_\phi(r_{\tau} \mid r_{<t}, x)
\right]
\label{eq:sft-loss}
\end{equation}
where $r_{\tau}$ are tokens of the synthetic stakeholder conversation and the rubric text.
After SFT, the parameters $\phi_0$ serve as initialization for the reinforcement phase.

\paragraph{Stage II: GSPO over Rubric Proposals.}
\label{sec:gspo}

In the reinforcement phase, the manager acts as a stochastic policy 
$\pi_M(R \mid x)$ that proposes candidate rubrics.  Training is performed episodically over the tasks 
$(x, \mathcal{O}[U^{\star}]) \in \mathcal{D}$ 
defined in Section~\ref{sec:training_data}, 
where each task $x$ serves as an environment for rollout-based optimization.  For each task $x$, $K$ rubrics $\{R_k\}_{k=1}^K$ are generated, 
each conditioning a worker model $\pi_W$ that produces an output 
$y_k \sim \pi_W(\cdot \mid x, R_k)$. 
The stakeholder utility $r_k = U^\star(y_k \mid x)$ serves as the scalar return. 
We compute a group-normalized baseline 
$\bar{r} = \tfrac{1}{K}\sum_k r_k$ 
and standardized advantages 
$\hat{A}_k = (r_k - \bar{r}) / \mathrm{std}(\{r_j\}_{j=1}^K)$.

\smallskip
\noindent
The decomposer parameters are updated to maximize the \textit{Group Sequence Policy Optimization (GSPO)} objective,
which replaces the token-level importance ratios of GRPO~\cite{shao2024deepseekmathpushinglimitsmathematical} 
with a length-normalized, sequence-level ratio $s_k(\phi)$:
\begin{align}
\mathcal{J}_{\mathrm{GSPO}}(\phi)
= \mathbb{E}_{x}\Bigg[
    &\frac{1}{K}\!\sum_{k=1}^{K}
      \min\!\bigg(
        s_k(\phi)\,\hat{A}_k,\;
        \nonumber \\ & \qquad
        \mathrm{clip}\!\big(s_k(\phi),\,1{-}\epsilon,\,1{+}\epsilon\big)\hat{A}_k
      \bigg)
    \nonumber\\
    &-\ \beta\, D_{\mathrm{KL}}\!\big(\pi_\phi(\cdot\mid x)\,\|\,\pi_{\mathrm{ref}}(\cdot\mid x)\big)
    \nonumber\\
    &-\ C_{\mathrm{clarify}}(q_{1:T})
    \;-\ C_{\mathrm{compute}}(R_k)
\Bigg]
\label{eq:lgsfo}
\end{align}

\begin{align}
s_k(\phi)
&=\left(
\frac{\pi_\phi(R_k \mid x)}{\pi_{\mathrm{old}}(R_k \mid x)}
\right)^{1/|R_k|}
\nonumber\\
&=\exp\!\Bigg(
\frac{1}{|R_k|}\sum_{t=1}^{|R_k|}
\log
\frac{\pi_\phi(z_{k,t}\mid x, z_{k,<t})}
     {\pi_{\mathrm{old}}(z_{k,t}\mid x, z_{k,<t})}
\Bigg)
\label{eq:sk}
\end{align}

\smallskip
\noindent
where \(z_{k,t}\) denotes the \(t\)-th token in the rubric sequence 
\(R_k = (z_{k,1}, \dots, z_{k,|R_k|})\), generated autoregressively by 
the manager policy \(\pi_\phi\) given the task context \(x\) and the 
preceding tokens \(z_{k,<t}\).

The KL term enforces a trust region around the reference policy 
$\pi_{\mathrm{ref}}$, preventing divergence during training. 
At this stage, the auxiliary regularizers $C_{\mathrm{clarify}}$ and 
$C_{\mathrm{compute}}$ are derived from metadata collected during rubric generation.

\textbf{Clarification cost.}
$C_{\mathrm{clarify}}(q_{1:T})$ depends on the number and difficulty of clarification 
questions during the stakeholder–manager dialogue, following a logarithmic 
normalization to model diminishing burden:
\begin{equation}
C_{\mathrm{clarify}}(q_{1:T})
= \frac{\log(1 + 0.1n_{\text{easy}} + 0.5n_{\text{med}} + n_{\text{hard}})}{\log(16)},
\end{equation}
where $n_\text{easy}, n_\text{med}, n_\text{hard}$ denote the numbers of clarification questions that the stakeholder agent classifies—via a fixed LLM prompt specifying criteria for easy, medium, and hard difficulty—into the corresponding cognitive-burden levels.
This discourages excessive or complex clarifications while tolerating minor ones.

\textbf{Compute cost.}
$C_{\mathrm{compute}}(R_k)$ measures the resource cost of rubric verification as a 
weighted combination of monetary and time costs:
\begin{equation} \label{eq:cost}
C_{\mathrm{compute}}(R_k)
= \frac{\log(1 + w_\text{cost} c_{\text{usd}} + w_\text{time} t_{\text{sec}})}{\log(1 + \alpha)},
\end{equation}
where $w_\text{cost} c_{\text{usd}}$ and $w_\text{time} t_{\text{sec}}$ denote weighted LLM verification cost and  wall-clock time. Both are fitted to logarithmic curves for smooth saturation 
across task sizes (weights specified in Appendix \ref{appendix:reproducibility}). To further stabilize learning and improve sample efficiency, we extend GSPO with \emph{prioritized experience replay}.  After each rollout, we record the mean return of each episode and for an iterative epoch we selectively replay episodes with mean policy returns in the bottom $N$-th percentile across eposides in training epoch. This emphasizes failure cases where the decomposer underperformed, allowing the policy to learn corrective refinements to low-utility rubric proposals.
This modification follows similar principles to prioritized sampling in reinforcement learning~\cite{schaul2016prioritizedexperiencereplay}, but applied at the rubric-group level rather than individual state transitions.

The final training loss is 
$\mathcal{L}(\phi) = -\,\mathcal{J}_{\mathrm{GSPO}}(\phi)$. Pseudo code for the full procedure is given in Algorithm~\ref{alg:GSPO}.
Once trained, the decomposer can be used to steer generation at test time
via rubric-conditioned sampling, as described next.

\begin{algorithm}[!t]
\caption{Group Sequence Policy Optimization (GSPO) for Rubric Generation}
\label{alg:GSPO}
\small
\begin{algorithmic}[1]
\Require initial decomposer parameters $\phi$ (from SFT), worker policy $\pi_W$, stakeholder simulator $U^\star$, 
dataset $\mathcal{D}$ of tasks, group size $K$, epochs $E$, clip $\epsilon$, KL coefficient $\beta$, learning rate $\eta$, 
max dialogue turns $T$, replay buffer $\mathcal{B}$, replay percentile $p$

\State $\phi_{\mathrm{ref}} \gets \phi$

\For{epoch $= 1, \ldots, E$}
  \For{$x \in \mathcal{D}$}
    \State Initialize dialogue history $h_0 \gets \emptyset$ 
    \For{$t = 1, \ldots, T$}
      \State Propose rubric draft $R_t \gets \mathfrak{D}_\phi(x, h_t)$
      \State Get stakeholder feedback $\delta_t \sim U^\star(\cdot \mid R_t, x)$
      \State Update history $h_{t+1} \gets h_t \cup \{(R_t, \delta_t)\}$
      \If{feedback is satisfactory}
          \State \textbf{break}
      \EndIf
    \EndFor

    \State Sample $K$ rubric proposals: $\{R_k\}_{k=1}^K \sim \pi_\phi(\cdot \mid x, h_t)$

    \For{$k = 1, \ldots, K$}
      \State Generate worker output $y_k \sim \pi_W(\cdot \mid x, R_k)$
      \State $r_k \gets U^\star(y_k \mid x)
          - \lambda_{\mathrm{clarify}}\,C_{\mathrm{clarify}}(h_t)
          - \lambda_{\mathrm{compute}}\,C_{\mathrm{compute}}(R_k)$
    \EndFor

    \State $\bar{r} \gets \frac{1}{K}\sum_{k=1}^K r_k$
    \State $\hat{A}_k \gets (r_k - \bar{r}) / \mathrm{std}(\{r_j\}_{j=1}^K)$

    \For{$k = 1, \ldots, K$}
      \State Compute importance weight $s_k(\phi)$ (Eq.~\ref{eq:sk})
      \State Compute clipped policy loss $\mathcal{L}_k$ (Eq.~\ref{eq:lgsfo})
    \EndFor
    \State $\mathcal{L} \gets \frac{1}{K}\sum_{k=1}^K \mathcal{L}_k$
    \State Store $(x, \{R_k\}_{k=1}^K, \mathcal{L}, \bar{r})$ in buffer $\mathcal{B}$
  \EndFor

  \State // \textit{Prioritized experience replay:}
  \State Select bottom $p$-th percentile episodes by $\bar{r}$ from $\mathcal{B}$
  \For{each stored $(x, \{R_k\}, \mathcal{L}, \bar{r})$ in replay set}
      \State Update $\phi \gets \phi - \eta \nabla_\phi \mathcal{L}$
  \EndFor
\EndFor
\State Return trained decomposer parameters $\phi^\star \gets \phi$
\end{algorithmic}
\end{algorithm}

\subsection{Test-Time Steering via Rubric-Scaled Policies}
\label{sec:scaled-policy}

After training, the decomposer enables interpretable control over worker behavior 
without any additional gradient updates.
At test time, ARCANE steers worker agent behavior by conditioning worker policies on the generated rubric $R^\star$ as shared context.  
This defines a joint worker policy
\begin{equation}
\pi^{\text{coop}}_W(y \mid x, R^\star) = \prod_i \pi_{W_i}(y \mid x, R^\star),
\end{equation}
whose collective behavior is implicitly aligned toward the stakeholder’s true utility $U^*$.  Since $R^\star$ defines $\hat{u}_\phi$ as a monotonic proxy for the stakeholder’s true utility $U^*$, it can serve as an oracle for ranking candidate outputs.  

In practice, we can exploit this property to achieve test-time scaling through multiple strategies:
\begin{itemize}
\item \textbf{Best-of-$K$ sampling:} 
Generate $K$ candidates from $\pi^{\text{coop}}_W(y \mid x, R^\star)$,  score them via $\hat{u}_\phi$,  and select the highest-scoring one.
\item \textbf{Importance reweighting:} 
Resample candidates in proportion to their rubric scores such as performed by \cite{liu2025tisdpotokenlevelimportancesampling}.
\item \textbf{Tree or beam search:} 
Use rubric scores as heuristics to guide structured decoding  (e.g., in code or reasoning tasks).
\end{itemize}

Because rubrics are interpretable, stakeholders can directly 
inspect, edit, or override the criteria $\{c_j\}$ and weights $\{w_j\}$ used at inference time. 
This provides a transparent mechanism for configurable alignment, 
linking training-time optimization to post-hoc human control.

\section{Experiments}\label{sec:experiments}

Our experiments aim to answer three research questions:
\begin{itemize}
    \item \textbf{RQ1 (Usefulness):} Does guiding the worker team with learned rubrics increase the utility of their outputs versus unguided generation?
    \item \textbf{RQ2 (Faithfulness):} Does the manager agent learn to rank candidates at test time in a manner consistent with the rankings computed from the stakeholder utility $U^\star$?
    \item \textbf{RQ3 (Interpretability \& Efficiency):}  Are the rubrics generated by the manager agent compact, legible, and practical to verify?
\end{itemize}

\subsection{Dataset and Domain}
We use the \textbf{GDPVal corpus}~\cite{patwardhan2025gdpvalevaluatingaimodel}, a large-scale benchmark designed for evaluating goal-directed agents that operate over challenging real-world tasks. 
Each task in GDPVal represents a multi-step work episode requiring agents to manipulate, analyze, or synthesize structured files—such as spreadsheets, documents, and figures.
This setting reflects practical domains like data analysis, document authoring, or visual reporting, where correctness must be demonstrated through verifiable outputs rather than language-only responses.

After filtering out tasks with unsupported file types (e.g., audio, PSD, CAD, or Apple Pages), we retain \textbf{219 tasks}, partitioned into \textbf{175 train} and \textbf{44 evaluation} episodes. 
All resources are file-based: workers must emit artifacts such as \texttt{.md}, \texttt{.xlsx}/\texttt{.csv}, \texttt{.pdf}/\texttt{.docx}, or \texttt{.png}/\texttt{.jpg}, enabling deterministic verification by downstream evaluators.

\subsection{Rubric Design and Labeling Stages}
Each task is paired with a staged \textbf{gold rubric} $R^\star$, written by expert annotators to cover three complementary aspects of task quality:
\begin{itemize}
    \item \textbf{Stage 1 – Gate:} Checks basic structure and task compliance (e.g., “Is the required file present?”, “Does the output follow the expected schema?”).
    \item \textbf{Stage 2 – Verification:} Validates factual or logical correctness, often mixing code-based tests with LLM judges to confirm that computations, references, and data align with the task description.
    \item \textbf{Stage 3 – Quality:} Assesses higher-level attributes, e.g. clarity, completeness, style, and usefulness to end user.
\end{itemize}
This staged decomposition was given as guidance to human rubric authors, ensuring that each gold rubric collectively covered structural, semantic, and qualitative dimensions of “goodness” for the task. Each rubric targets 9–12 criteria, distributed by weight across stages (\emph{Gate} 20–30\%, \emph{Verification} 40–50\%, \emph{Quality} 20–30\%). 
Criteria are implemented as either rule-based checks or LLM judgments. 
Overall, the collection includes 2,601 criteria: 1,931 LLM judges (74.2\%) and 670 code rules (25.8\%).

\subsection{Rubric Evaluation Mechanics}
Each rubric criterion is evaluated by either a vision-language model (VLM) judge or a deterministic code rule.
LLM-judge rules use a fixed multimodal model snapshot capable of reading rendered artifacts (e.g., spreadsheets, PDFs, figures) and apply templated 0–1 scoring prompts. To minimize verifier error, we (i) use verifier models from different model families as the workers they judge to avoid "self enforcement bias" \cite{huang2025empiricalstudyllmasajudgellm}, and design rules to be easily verifiable, exploiting "asymmetry of verification" \cite{huang2025reinforcementlearningrubricanchors}. Code rules are executed as Python functions with access to all intermediate and final task resources, allowing deterministic checks such as schema validation, bounds tests, or consistency verification. 
This hybrid evaluation enables reproducible low-level validation while reserving LLM judges for higher-level reasoning and qualitative assessment.

\paragraph{Baselines.} 
We compare four baseline configurations, all using identical worker model configurations:
\begin{itemize}
    \item \textbf{Best-of-N (No Rubric):} 
    A baseline in which the manager agent performs no rubric generation or stakeholder interaction. 
    Workers directly attempt the task based on its description $x$, and after all completions, the manager ranks the resulting $N$ outputs $\{y\}$ using its own subjective preference model $\hat{u}$.
    
    \item \textbf{Rubric Generation (SFT):} 
    The manager, finetuned using the SFT model from Section~\ref{sec:sft}, first interacts with the stakeholder to elicit and generate a rubric, which is then distributed to $N$ parallel workers for best-of-N rollouts. 
    The resulting outputs $\{y\}$ are subsequently scored by the learned utility estimator $\hat{u}_{\phi_{\text{SFT}}}$.
    
    \item \textbf{Rubric Generation (GSPO):} 
    Identical to the previous setup, but using the GSPO-trained manager from Section~\ref{sec:gspo}. 
    This isolates the effect of preference-optimized rubric generation compared to supervised finetuning.
    
    \item \textbf{Oracle Rubric:} 
    A privileged best-of-N configuration in which the manager is given direct access to the ground-truth rubric $R^\star$.
    The manager distributes $R^\star$ to workers to guide task execution, and final outputs are ranked according to the corresponding ground-truth utility $U^\star$.
    This serves as an empirical upper bound on the achievable gains from rubric guidance.

\end{itemize}

The sole difference across baselines is the source of rubric guidance (none, SFT-generated, GSPO-generated, or oracle). This controlled setup isolates the impact of rubric quality on worker performance. Training details can be found in Appendix \ref{appendix:reproducibility} and worker parameterizations in Appendix \ref{appendix:tools}.

\subsection{Results}\label{sec:exp:results}

\subsubsection{Usefulness (RQ1)}\label{sec:results:usefulness}

We report average true return $U^*(y_i \mid x_i)$ across 44 held-out tasks $x_i$.

\begin{figure}[t]
    \centering
    \includegraphics[width=0.48\textwidth]{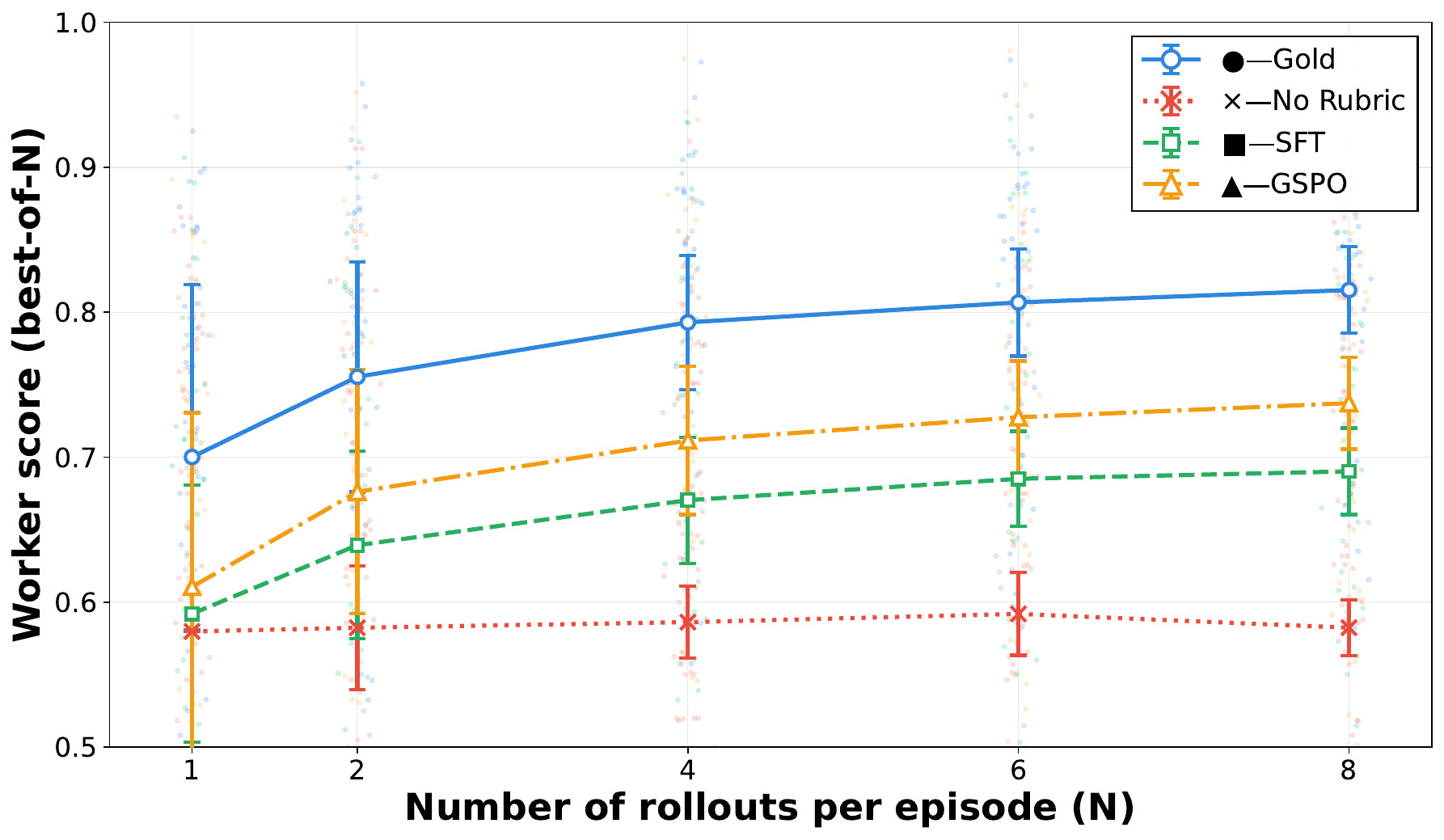}
    \caption{
        \textbf{Test-Time Compute Scaling: SFT vs GSPO vs No Rubric and Oracle Baselines.}
        Mean returns on the 44-task evaluation set under single-sample generation ($N{=}1$)
        and best-of-$N$ sampling ($N{=}1{\ldots}8$).
        Error bars show 95\% bootstrap confidence intervals.
        The Oracle (Gold Rubric) baseline is an upper bound on rubric guidance effectiveness.
    }
    \label{fig:compute-scaling}
\end{figure}

\textbf{Findings.}
Figure~\ref{fig:compute-scaling} shows mean returns under increasing test-time compute
(best-of-$N$ sampling, $N{=}1{\ldots}8$).
Guided variants consistently outperform the unguided baseline.
As shown in Table \ref{tab:usefulness},  mean return rises modestly from $0.58$ (No Rubric) to $0.62$ (GSPO) at $N=1$,
while the Oracle Rubric reaches $0.70$.
At $N{=}8$, the ordering remains (No Rubric $<$ SFT $<$ GSPO $<$ Gold) with larger increase in means
$0.58$, $0.68$, $0.74$, and $0.81$.
This monotonic improvement shows that learned rubrics effectively encode stakeholder preferences,
boosting task-level return without modifying the worker model or decoding parameters.

A one-sided Wilcoxon signed-rank test on the evaluation tasks confirms statistical significance of GSPO's mean returns over SFT returns ($p{=}0.0182$ at N=8), with mean per-task improvement $+0.044$ ($W^+{=}592$, $W^-{=}269$; 22 episodes where GSPO wins vs.\ 19 where SFT wins). At best-of-eight sampling, mean returns are $0.6817$ (SFT) vs.\ $0.7460$ (GSPO), confirming a statistically significant margin at $\alpha{=}0.05$. Reinforcement fine-tuning thus improves rubric utility beyond supervised learning alone. All guided models (SFT, GSPO, and Oracle) exhibit nearly identical best-of-$N$ scaling slopes:
each doubling of $N$ yields an average $\approx\!+0.03$ absolute gain. Although their absolute returns differ, the parallel scaling curves between the learned (SFT, GSPO) and Oracle conditions indicate that the learned rubrics confer a similarly consistent advantage per additional sample. In other words, once a rubric is introduced, stochastic exploration improves quality at nearly the same rate as if evaluation were performed by the oracle critic—suggesting that learned rubrics approximate its scoring function even without direct access to $U^\star$.

\begin{table}[t]
\centering
\small
\setlength{\tabcolsep}{6pt} 
\resizebox{\linewidth}{!}{
\begin{tabular}{lcc}
\toprule
\textbf{Model} & \textbf{Mean$\!\pm\!$sd ($N{=}1$)} & \textbf{Mean$\!\pm\!$sd ($N{=}8$)} \\
\midrule
No Rubric (LLM-Judged) & $0.58 \pm 0.01$ & $0.58 \pm 0.02$ \\
SFT Model              & $0.59 \pm 0.09$ & $0.68 \pm 0.03$ \\
GSPO Model             & $0.62 \pm 0.12$ & $0.74 \pm 0.03$ \\
Oracle Rubric (Baseline) & $0.70 \pm 0.12$ & $0.81 \pm 0.03$ \\
\bottomrule
\end{tabular}
}
\caption{Mean returns $\pm$ \emph{standard deviation} across evaluation tasks under single-sample ($N{=}1$) and best-of-eight ($N{=}8$) generation.}
\label{tab:usefulness}
\end{table}

\subsubsection{Faithfulness (RQ2)}\label{sec:results:faithfulness}

We measure agreement between rubric-induced scores $\hat{u}_\phi$
and oracle utilities $U^\star$ using Normalized Discounted Cumulative Gain across 8 worker rollouts per episode (NDCG@8),
averaged across 44 evaluation tasks. For each task, we evaluate the same set of eight rollouts but rank them once under the learned rubric $\hat{u}_\phi$ and once under the oracle $U^\star$.
Formally, $\mathrm{NDCG@8}=\Big(\sum_{i=1}^8 \mathrm{rel}_i/\log_2(i+1)\Big)\Big/\Big(\sum_{i=1}^8 \mathrm{rel}^\star_i/\log_2(i+1)\Big)$

where $\mathrm{rel}_i$ is the oracle relevance (i.e., $U^\star$ score)
of the item ranked at position~$i$ by $\hat{u}_\phi$,
and $\mathrm{rel}^\star_i$ denotes the oracle relevance under the ideal (descending) ordering.
NDCG quantifies listwise consistency—rewarding rubrics that place
high-utility candidates near the top while discounting errors lower in the list
\cite{zhou2024drpo,zhao2025ppa}. Each system generates eight candidates using its own rubric guidance. We then take these generated candidates and evaluate them under both the system’s rubric ($\hat{u}_\phi$)
and the oracle utility function ($U^\star$),
measuring how faithfully the rubric’s relative preferences track true stakeholder utility
on identical candidate sets.

We also include a \textbf{No-Conversation (Base Rubric)} baseline, in which the manager still generates a rubric but does so using the base model \textit{without any stakeholder dialogue}. This variant isolates the model’s intrinsic understanding of the stakeholders preference for each task by taking it directly from the description $x$, independent of explicit preference elicitation, serving as a lower-bound comparison for stakeholder-guided rubric learning.

\textbf{Results.}
Table~\ref{tab:faithfulness} shows steady gains:
NDCG@8 improves from 0.7998 (Base) to 0.8103 (SFT) and 0.8722 (GSPO),
approaching oracle alignment while maintaining similar variance.
The GSPO improvement corresponds to +8.3\% Precision@3
and a reduction in average rank swaps among top candidates
(2.20~$\rightarrow$~2.13 per task). No system achieved perfect top-3 agreement, underscoring the difficulty of fine-grained ranking even when mean utility is well aligned.

\begin{table}[h]
\centering
\small
\begin{tabular}{lcc}
\toprule
\textbf{Model--Rubric Pair} & \textbf{Mean NDCG@8} & \textbf{Std.} \\
\midrule
No-Conv (Base) & 0.7998 & 0.0807 \\
SFT Rubric     & 0.8103 & 0.0881 \\
GSPO Rubric    & \textbf{0.8722} & 0.0985 \\
\bottomrule
\end{tabular}
\caption{
Mean $\pm$ standard deviation of NDCG@8 between
rubric-induced and oracle rankings ($U^\star$).
Higher is better.
}
\label{tab:faithfulness}
\end{table}

\textbf{Domain trends.}
Improvements are strongest for subjective, language-heavy domains: content/communication (+11.5\%, NDCG = 0.905) and legal/compliance (+12.5\%, 0.848),
while operational tasks slightly regress (–8.1\%, 0.830)
and data-analysis tasks remain flat (+1.9\%, 0.822).
We conjecture that reinforcement fine-tuning excels where quality is multi-dimensional and hard to specify a priori—%
in content and legal work, rubrics capture nuanced aspects (tone, completeness, regulatory alignment)
that benefit from learned discriminative features.
Thus, reinforcement fine-tuning is most valuable when quality is subjective and decomposable.

\subsubsection{Interpretability (RQ3)}\label{sec:results:interpretability}

Are learned rubrics compact, legible, and auditable by practitioners?

\textbf{Metrics.}
We assess (1)~number of generated criteria, (2)~tokens per criterion, and (3)~criteria weight distributions—the variance of point allocation across criteria in the Gate, Verification, and Quality stages,
where higher entropy implies more balanced structure. We supplement quantitative measures with qualitative comparison
of the Oracle and GSPO rubrics in Appendix \ref{app:rubric-comparison}.

\textbf{Results.}
Table~\ref{tab:rubric-structure-summary} shows that SFT and GSPO rubrics nearly match the Oracle standard: $\sim$12 criteria per rubric and 17–18 tokens per description. GSPO rubrics are marginally shorter (–1 token, –5.5 \%), allocate +14 \% more weight value to criteria specifically measuring quality, and exhibit lower variance (3.5 vs 4.2). These patterns indicate that reinforcement fine-tuning yields
more consistent phrasing and stage balance without verbosity or collapse. Appendix~\ref{app:rubric-comparisondD1} compares representative criteria from the
\textit{Government – Recreation Workers} task.
GSPO rubrics retain the structure of the Oracle while adopting
shorter, imperative phrasing
(e.g., “detects duplicates,” “confirms preferences handled”)
and stable point distributions
(Gate $\approx$ 6\,\%, Verification $\approx$ 52\,\%, Quality $\approx$ 42\,\%).
Fine-tuning therefore improves linguistic regularity
without distorting semantics.

We can see on inspection that GSPO produces rubrics with evaluation criteria that remain transparent, auditable,
and faithful to human intent. Together with RQ1–RQ2, these results demonstrate that learned rubrics are
\emph{useful}, \emph{faithful}, and \emph{interpretable}.

\section{Conclusion}
We presented ARCANE, a rubric-based framework for interpretable and configurable alignment of agents to stakeholder preferences. By explicitly modeling latent stakeholder preferences as structured rubrics and training a manager agent to elicit and synthesize rubrics through dialogue with the stakeholder, ARCANE provides both theoretical grounding (Appendix \ref{app:operator}) and practical method for test-time alignment. This enables steerability without retraining and improves utility on complex tasks requiring multi-step reasoning. Our experimental results indicate that learned rubrics can reliably guide worker policies, approximate oracle ranking, and remain compact and auditable.

\paragraph{Limitations.} Our experiments evaluate a manager coordinating a single worker; while the framework extends to multiple workers (Section \ref{sec:background}), we have not validated multi-worker coordination dynamics. GDPVal tasks, though requiring multi-step tool use, are discrete episodes—stronger long-horizon evidence would require evaluation on extended deployments, such as in complex multi-agent workflows requiring manager orchestration \cite{masters2025orchestratinghumanaiteamsmanager}. Our No Rubric baseline uses an RLHF-trained worker (Qwen-3), demonstrating that training-time alignment alone does not provide test-time adaptability; however, direct comparison to alternative test-time methods such as Generative Reward Models would help to characterize tradeoffs between interpretability and performance. Finally, GSPO's reward shaping is flexible (we include terms for cost and latency), but we lack structural regularizers against spurious criteria. The manager agent could learn criteria that correlate with utility without being causally meaningful; incorporating causal or invariance penalties into rubric learning could improve the faithfulness to true preferences.

\paragraph{Future Work.} Promising directions include: \textit{(1)} dynamic rubric maintenance through adaptive revision during execution and bidirectional feedback from workers flagging ambiguous or conflicting criteria; \textit{(2)} systematic comparison with alternative test-time alignment methods (e.g., GenRM, GRAM) and extension to multi-agent settings with multiple coordinating workers; and \textit{(3)} scaling to large-scale heterogeneous preference data to improve coverage and generalization on long-tail domains.

\bibliography{aaai2026}
-----------------------
\appendix

\section{Operator Calibration of Supervision (On-Support)}
\label{app:operator}

In our experiments, we show the efficacy of our reinforcement fine-tuning procedure for learning rubrics from gold rubrics. In this Appendix we show that no matter how the stakeholder utility is observed (pointwise scores, pairwise wins, listwise picks), if we (i) define the right \emph{target} for that supervision, (ii) train with a \emph{strictly proper} loss,
and (iii) stay on the same decoding support we use at test time, then the learned scorer $g_\phi$ matches
the supervision target on that support and therefore induces the same ordering as the latent utility $U^*$.

\paragraph{Setup}
Let $U^*(y\mid x)$ be the latent stakeholder utility. At inference we draw a candidate \emph{multiset}
$C\sim\mathcal{C}(x)$; its support $S=\mathrm{supp}(\mathcal{C})$ is the \emph{decoding support} (where we actually look).
An \emph{observation operator} $\mathcal{O}(U^*)$ is a stochastic mechanism emitting a supervision signal
$Z$ given an \emph{instance} $\mathcal{I}$ (the information the operator conditions on).
Examples of $\mathcal{I}$ for different types of obvervations would be: $\mathcal{I}=(x,p,y)$ for pointwise exemplar policies, $(x,y_i,y_j)$ for pairwise preferences, and $(x,C)$ for listwise preferences.

The model maps instances to a score $g_\phi(\mathcal{I})$, then applies a link $h$ (identity / logistic / softmax) to produce a prediction $T_\phi(\mathcal{I})=h(g_\phi(\mathcal{I}))$ that is compared to a \emph{target functional}
$T^*(U^*)(\mathcal{I})$ via a strictly proper loss $L$.

\begin{table*}[t]
\centering
\caption{Example supervision operators, targets, links, and strictly proper losses.}
\label{tab:ops-targets}
\begin{tabular}{@{}l l l l l l@{}}
    \toprule
    \textbf{Supervision} & \textbf{Instance $\mathcal{I}$} & \textbf{Obs. $Z$} & \textbf{Target $T^*(U^*)$} & \textbf{Link $h(g_\phi)$} & \textbf{Loss $L$} \\
    \midrule
    Pointwise & $(x,y)$ & $r$ &
    $\mathbb{E}[r\,|\,x,y]$ &
    $g_\phi(y)$ & MSE \\
    Pairwise & $(x,y_i,y_j)$ & $\mathbf{1}[y_i\!\succ\! y_j]$ &
    $\sigma\!\big(\beta(U^*(y_i){-}U^*(y_j))\big)$ &
    $\sigma\!\big(g_\phi(y_i){-}g_\phi(y_j)\big)$ & Bernoulli NLL \\
    Listwise & $(x,C)$ & $y^\star\!\in C$ &
    $q^*(y\!\mid\!C)\propto e^{\beta U^*(y)}$ &
    $\mathrm{softmax}\!\big(g_\phi(y)/\tau\big)$ & $-\log t(y^\star)$ \\
    \bottomrule
\end{tabular}
\end{table*}

\paragraph{Assumptions.}
\textbf{(A1) Coverage.} Training and decoding use the same sampler; guarantees hold a.s.\ on $S=\mathrm{supp}(\mathcal{C})$.\\

\textbf{(A2) Composite realizability (on $S$).} There exists $\phi_0$ with $h(g_{\phi_0}(\mathcal{I}))=T^*(U^*)(\mathcal{I})$ a.s.\ for $\mathcal{I}\in S$.
\\
\textbf{(A3) Properness.} $L$ is strictly proper for $(\mathcal{O},T^*)$: for each $\mathcal{I}$, the conditional Bayes risk
$t\mapsto \mathrm{E}[L(Z,t)\mid \mathcal{I}]$ is uniquely minimized at $t=T^*(U^*)(\mathcal{I})$.

\paragraph{Theorem (Operator calibration on the decoding support).}
Let $\mathcal{R}(\phi)=\mathrm{E}\big[L\big(Z,\,h(g_\phi(\mathcal{I}))\big)\big]$ be the population risk.
Under (A1)–(A3), any minimizer $\phi^\star\in\arg\min_\phi \mathcal{R}(\phi)$ satisfies
\[
h\!\left(g_{\phi^\star}(\mathcal{I})\right)\;=\;T^*(U^*)(\mathcal{I})
\qquad\text{for a.s.\ }\mathcal{I}\in S.
\]
In particular, $g_{\phi^\star}$ and $U^*$ induce the same order on $S$ whenever $h$ and $T^*$ are strictly monotone in the relevant score/utility differences (identity, logistic, softmax). Examples of such cases are given in Table \ref{tab:ops-targets} and later Corollaries.

\emph{Proof sketch}
\textbf{(1) Bayes risk.} We Define $R_\phi(\mathcal{I})=\mathrm{E}[L(Z,\,h(g_\phi(\mathcal{I})))\mid \mathcal{I}]$ and
$R^*(\mathcal{I})=\inf_t \mathrm{E}[L(Z,t)\mid \mathcal{I}]$. Strict properness gives
$R_\phi(\mathcal{I})\ge R^*(\mathcal{I})$ with equality iff $h(g_\phi(\mathcal{I}))=T^*(U^*)(\mathcal{I})$.\\

\textbf{(2) Pointwise $\Rightarrow$ global.} Taking expectations over $\mathcal{I}\sim S$,
$\mathcal{R}(\phi)-\mathrm{E}[R^*(\mathcal{I})]=\mathrm{E}[R_\phi(\mathcal{I})-R^*(\mathcal{I})]\ge 0$,
with equality iff the pointwise equality holds a.s.

\textbf{(3) Realizability.} By (A2) there exists $\phi_0$ attaining equality, hence any global minimizer $\phi^\star$
must also achieve equality a.s.\ on $S$. \qedsymbol

\paragraph{Corollary (Listwise / InfoNCE).}
For listwise $\mathcal{I}=(x,C)$ with oracle $q^*(y\mid C)\propto \exp\{\beta U^*(y)\}$,
link $q_\phi(y\mid C)=\mathrm{softmax}(g_\phi(y)/\tau)$, and log loss,
\[
\mathrm{E}[L]\;=\;\mathrm{E}\!\big[\,H(q^*,q_\phi)\,\big]
\;=\;\mathrm{E}\!\big[\,H(q^*)+D_{\mathrm{KL}}(q^*\|q_\phi)\,\big],
\]
so minimizing $\mathrm{E}[L]$ minimizes the KL divergence between oracle and model listwise distributions.
If $q_\phi=q^*$ on $C$, then $g_{\phi^\star}(y)-g_{\phi^\star}(y')=\tau\beta\,[U^*(y)-U^*(y')]$ for all $y,y'\in C$:
scores are identifiable up to per-group positive scale and additive shift (which do not affect selection).

\paragraph{Corollary (Pairwise / logistic).}
For $\mathcal{I}=(x,y_i,y_j)$ with target $T^*_{\mathrm{pw}}=\sigma(\beta(U^*(y_i)-U^*(y_j)))$,
link $h(g_\phi)=\sigma(g_\phi(y_i)-g_\phi(y_j))$, and Bernoulli log-loss,
any risk minimizer satisfies
$\sigma(g_{\phi^\star}(y_i)-g_{\phi^\star}(y_j))=T^*_{\mathrm{pw}}$ a.s.\ on $S$.
Hence $\mathrm{sign}\!\big(g_{\phi^\star}(y_i)-g_{\phi^\star}(y_j)\big)=\mathrm{sign}\!\big(U^*(y_i)-U^*(y_j)\big)$ on $S$
(order equivalence; only differences matter).

\paragraph{Notes and scope.}
(i) \textbf{On-support guarantee.} All statements hold on the decoding support $S$ induced by your sampler; changing $K$, $\tau$, or the sampler at test time alters $S$ and can change behavior.\\
(ii) \textbf{Temperature/scale.} With softmax, utilities are identifiable only up to affine transforms within a group; selection and ranking are invariant to these transforms.\\
(iii) \textbf{Bandit/off-policy.} With logged propensities, inverse-propensity or doubly-robust estimators consistently estimate $\mathcal{R}(\phi)$ under standard overlap; without overlap, unbiased estimation is impossible.

\paragraph{Connection to GSPO}
This theorem calibrates the \emph{evaluator} $g_\phi$ via a strictly proper loss so that it orders candidates like $U^*$ on $S$. Separately, the \emph{manager} (rubric generator) is trained with GSPO on returns $U^*-\lambda C$
under a trust region $KL(\pi_{\mathrm{mgr,new}}\|\pi_{\mathrm{mgr,ref}})\!\le\!\delta$ over rubric token sequences;
PPO-style arguments yield non-decreasing expected return up to $O(\delta+\epsilon^2)$.

\section{Gold Rubric Construction}\label{appendix:rubric-taxonomy}

Gold rubrics $R^\star$ consist of three stages: \textbf{Gate} (20--30\%, checking structure/completeness), \textbf{Verification} (40--50\%, validating correctness), and \textbf{Quality} (20--30\%, assessing polish and stakeholder alignment). Each rubric contains 9--12 criteria total.

\textbf{Criterion design principles:} (1) \emph{Specificity}---avoid vague criteria like ``output is good''; instead specify observable properties (e.g., ``report contains 3--5 recommendations''). (2) \emph{Verifiability}---each criterion is checkable via code or LLM judge with high reliability. (3) \emph{Independence}---minimize redundancy across criteria. (4) \emph{Partial credit}---allow graduated scoring where appropriate.

\textbf{Verifier types:} Code-based verifiers handle deterministic checks (file format, field presence, arithmetic). LLM judges (GPT-4o, temperature 0.3) handle semantic criteria (tone, factual consistency, argument quality). We prefer code when possible for objectivity and cost.

\textbf{Example rubric (Healthcare Financial Report):}
\begin{itemize}
    \item \textbf{Gate (25 pts):} Valid DOCX (10); required sections present (10); includes chart/table (5). All code-based.
    \item \textbf{Verification (50 pts):} Revenue totals match source (20, code); figures internally consistent (15, LLM+code); correctly identifies top-3 revenue/expense categories (15, LLM).
    \item \textbf{Quality (25 pts):} Professional tone (10, LLM); actionable recommendations (10, LLM); polished formatting (5, LLM).
\end{itemize}

\textbf{Inter-annotator reliability (20 dual-annotated tasks):} Cohen's $\kappa = 0.82$ on criterion inclusion; mean absolute difference of 3.2 points on weights; 91\% agreement on verifier type selection. Disagreements resolved by senior annotator adjudication.

\section{Worker Descriptions}\label{appendix:tools}
 All workers were run using GPT-5 (check pointed as of 25/10/2025, tool access, and decoding parameters temperature 0.7, seed 42). Workers have access to 40 tools organized into 8 functional categories. All file operations are scoped to task-specific directories; workers cannot access external APIs or network resources. A definitive list of Tools can be found in Table 4.

\begin{table}[h]
\centering
\small
\caption{Worker tool categories and capabilities.}
\label{tab:tools}
\begin{tabular}{@{}lcp{5.5cm}@{}}
\toprule
\textbf{Category} & \textbf{\#} & \textbf{Description} \\
\midrule
File I/O & 9 & \texttt{read/write\_file}, \texttt{read/write\_xlsx}, \texttt{read/write\_docx}, \texttt{read/write\_pdf}, \texttt{list\_files}. Excel via pandas; Word/PDF text extraction and generation. \\
\addlinespace[0.5em]
Python Exec & 2 & \texttt{python\_exec(code)}, \texttt{python\_eval(expr)}. Sandboxed with \texttt{pandas}, \texttt{numpy}, \texttt{matplotlib}, \texttt{scipy}, \texttt{scikit-learn}, \texttt{seaborn}, \texttt{plotly}. Network disabled; 60s timeout. \\
\addlinespace[0.5em]
Data Analysis & 6 & \texttt{aggregate}, \texttt{filter\_data}, \texttt{sort\_data}, \texttt{merge\_data}, \texttt{pivot\_table}, \texttt{describe\_statistics}. Operate on pandas DataFrames. \\
\addlinespace[0.5em]
Visualization & 6 & \texttt{plot\_line}, \texttt{plot\_bar}, \texttt{plot\_scatter}, \texttt{plot\_histogram}, \texttt{plot\_pie}, \texttt{save\_plot}. Generate matplotlib/plotly charts. \\
\addlinespace[0.5em]
Doc Rendering & 3 & \texttt{render\_docx/pdf/xlsx\_to\_image}. Convert documents to images (150 DPI) for GPT-4o vision model input. \\
\addlinespace[0.5em]
Retrieval & 3 & \texttt{rag\_query}, \texttt{bm25\_query}, \texttt{embed\_text}. Dense and keyword-based search over task-provided corpora. \\
\addlinespace[0.5em]
OCR & 2 & \texttt{ocr\_image}, \texttt{ocr\_pdf}. Extract text from images and scanned PDFs. \\
\addlinespace[0.5em]
PDF Manip. & 4 & \texttt{pdf\_merge}, \texttt{pdf\_split}, \texttt{pdf\_extract\_pages}, \texttt{pdf\_add\_watermark}. \\
\addlinespace[0.5em]
Validation & 5 & \texttt{validate\_json}, \texttt{validate\_email}, \texttt{validate\_url}, \texttt{validate\_date}, \texttt{validate\_schema}. Check formats and constraints. \\
\bottomrule
\end{tabular}
\end{table}

\section{Rubric Interpretability Examples}
\label{app:rubric-comparison}

\subsection{Oracle vs GSPO Rubric Comparison}
\label{app:rubric-comparisondD1}

Table~\ref{tab:rubric-side-by-side} contrasts representative criteria from
the \textit{Government -- Recreation Workers} task.
Each row pairs a criterion from the human-authored gold rubric
with its nearest learned counterpart from the GSPO rubric within the same stage.
The comparison highlights how reinforcement fine-tuning preserved
structure and evaluability while producing more concise, imperative phrasing.

\begin{table*}[t]
\centering
\small
\begin{tabular}{@{}p{0.12\textwidth}p{0.41\textwidth}p{0.41\textwidth}@{}}
\toprule
\textbf{Stage} & \textbf{Gold Rubric Criterion (Reference)} & \textbf{GSPO Learned Criterion (Concise)} \\
\midrule
\textbf{Gate} &
\emph{``Verify the candidate produced a single Excel workbook with required sheets
and clearly labeled tables/columns that enable verification.''} &
\emph{``Checks workbook structure: single workbook, required sheets present, labeled tables, parsable for verification.''} \\[0.5em]
\textbf{Verification (Code)} &
\emph{``No table number should be assigned to more than one vendor across the layout.''} &
\emph{``Detects duplicate table assignments; each table ID unique across vendors.''} \\[0.5em]
\textbf{Verification (LLM)} &
\emph{``Location preferences, electricity needs, and adjacency requests are honored where feasible, or deviations are justified.''} &
\emph{``Confirms vendor preferences handled or justified; highlights unmet adjacency/electricity requests.''} \\[0.5em]
\textbf{Quality (LLM)} &
\emph{``Formatting, labeling, and layout are professional and accessible for city staff and vendors.''} &
\emph{``Assesses professionalism and accessibility of layout for staff and vendors.''} \\[0.5em]
\textbf{Quality (LLM)} &
\emph{``Evidence of contingencies for common risks (no-shows, power issues, last-minute changes).''} &
\emph{``Checks inclusion of contingency notes for expected operational risks.''} \\
\bottomrule
\end{tabular}
\caption{
Side-by-side comparison of representative criteria from the
\textit{Government -- Recreation Workers} task.
GSPO rubrics mirror gold structure and coverage but use shorter,
imperative phrasing (\textasciitilde5–6 tokens fewer on average),
improving legibility and auditability without loss of specificity.
}
\label{tab:rubric-side-by-side}
\end{table*}

\subsection{Rubric Structure Summary}

Table~\ref{tab:rubric-structure-summary} summarizes aggregate
weights and average token lengths per stage for the same task.

\begin{table}[H]
\centering
\small
\resizebox{\columnwidth}{!}{%
\begin{tabular}{@{}lccc@{}}
\toprule
\textbf{Stage} & \textbf{Gold Weight (\%)} & \textbf{GSPO Weight (\%)} & \textbf{Tokens/Rule (Gold $\rightarrow$ GSPO)} \\
\midrule
Gate & 5 & 6 & 22 $\rightarrow$ 18 \\
Verification & 53 & 52 & 26 $\rightarrow$ 19 \\
Quality & 42 & 42 & 16 $\rightarrow$ 14 \\
\bottomrule
\end{tabular}
}
\caption{
Per-stage weight and token-length comparison for the same task.
reinforcement fine-tuning maintains near-identical point distribution while shortening criteria
and regularizing phrasing across stages.
}
\label{tab:rubric-structure-summary}
\end{table}

\textbf{Observation.}
Across stages, GSPO preserves the same structural decomposition as the gold rubric,
retains measurable criteria, and avoids redundant wording.
The learned phrasing replaces passive descriptions
with active checks (``detects,'' ``confirms,'' ``assesses''),
reflecting consistent translation of human evaluation intent
into machine-auditable form.

\section{Reproducibility and Experimental Configuration}
\label{appendix:reproducibility}
Training with Group Sequence Policy Optimization (GSPO) described in Section~\ref{sec:method} ran for two epochs, each lasting approximately thirteen hours on a single NVIDIA~H100~(80\,GB) GPU. 
Each batch contained four rollouts, with a batch size of four and eight gradient-accumulation steps, giving an effective update batch size of 128 rollouts. 
All model variants shared the same worker policy, verifier design, and tool environment; only the source of rubric generation differed between baselines (none, SFT, GSPO, or oracle).

Training used the \texttt{Qwen3-8B} model with low-rank adaptation (LoRA rank $r=16$, $\alpha=16$). 
Roughly ninety-four million parameters ($0.67\%$ of the full model) were trainable. 
The worker model was \texttt{GPT--5} for all reported experiments, while \texttt{Claude-4.5-Haiku} was evaluated in limited runs and produced similar qualitative results, with main differences arising from API throughput constraints. 
The verifier large language model was \texttt{GPT--4.1}.  
All experiments were implemented using \texttt{PyTorch~2.13}, \texttt{TRL}, \texttt{Unsloth}, and \texttt{VLLM} (October~2025 releases).  
Mixed-precision training was performed in \texttt{BF16}, and NF4 quantization was applied for single-GPU training to reduce memory usage.  
Optimization employed \texttt{AdamW} with $\beta_1=0.9$, $\beta_2=0.99$, and a weight decay of~0.01.  
The learning rate was set to $5\times10^{-6}$ with a 10\% linear warm-up and cosine decay.

The dataset comprised a subset of the GDP-Val benchmark, with 175~training tasks (700~rollouts) and 44~held-out evaluation tasks.  
Evaluation used a hybrid verifier combining static code checks with LLM-judge criteria (both implemented with GPT-4.1).  
The primary evaluation metric was the maximum true return per worker within each rollout, $\max_i U^{\ast}(y_i)$.  
All experiments were run with a fixed random seed of~42, and reported values include 95\% bootstrap confidence intervals.

\paragraph{Supervised Fine-Tuning Baseline.}
Before GSPO training, we conducted a supervised fine-tuning (SFT) stage to establish a strong non-RL baseline. 
SFT was performed on \texttt{Qwen3-8B} (32k context) using LoRA with rank $r=16$ and $\alpha=16$, matching the adapter configuration used in GSPO.  
The learning rate was set to $2\times10^{-5}$, with a batch size of four per device and four gradient-accumulation steps, yielding an effective batch size of 16.  
One epochs of SFT was run with no gradient warmup.  The optimizer was \texttt{AdamW} with the same $\beta$-values and decay as GSPO.  
This SFT model served as the initialization for the SFT baseline reported in Section~\ref{sec:experiments}.

\paragraph{GSPO and Cost Penalties.}
All RL runs used $\beta=0.05$ for KL regularization, PPO clipping parameter $\epsilon=0.2$, and a group size of $K=4$.  
Two additional cost coefficients governed the reward shaping:
$\lambda_{\mathrm{cost}}=0.01$ (verifier compute cost) and 
$\lambda_{\mathrm{burden}}=0.05$ (stakeholder clarification burden), with 
\texttt{BURDEN\_SCALE}=15.0 for logarithmic saturation of clarification penalties.  
Under this scheme, one clarification incurred roughly an 18\% reward penalty, five clarifications 42\%, and ten clarifications 60\%.  

\paragraph{Cost-Time Weighting for Parallel Rubrics.}
The compute cost term $C_{\mathrm{compute}}(R_k)$ in Eq.~\ref{eq:cost} decomposes into two components: monetary cost (API expenses) and time cost (wall-clock execution time). 
To encourage rubrics that enable parallel execution and reduce latency, we weighted these components as $w_{\mathrm{cost}}=0.3$ and $w_{\mathrm{time}}=0.7$, prioritizing rubrics that minimize execution time over those that minimize API costs. 

\paragraph{Optimization Settings.}
A warmup fraction of 10\% was applied, the KL coefficient $\beta$ was tuned between $\{0.02, 0.05, 0.1\}$ in pilot runs, and $\beta=0.05$ provided the most stable learning.  
Batch size per device was one, with gradient accumulation of eight for an effective batch of~32.  
Each prompt generated $K=4$ rubric-conditioned rollouts per update.  

\paragraph{Compute and Environment.}
Single-GPU training consumed approximately twenty-six GPU-hours (two epochs at thirteen hours each), with average memory usage of 62\,GB using NF4 quantization.  
Experiments were run on Ubuntu~22.04~LTS with CUDA~12.2.  
Progress and replay buffers were tracked using Weights~\&~Biases, recording per-epoch returns and rollout statistics.

\paragraph{Determinism and Verification.}
To ensure reproducibility, all random seeds were fixed and deterministic \texttt{PyTorch} operations enabled.  
Ablation sweeps confirmed that $\lambda_{\mathrm{cost}}\!\in\!\{0,0.001,0.01,0.1,1.0\}$ and 
$\lambda_{\mathrm{burden}}\!\in\!\{0,0.01,0.05,0.1,0.2\}$ yield monotonic trade-offs between utility and compute burden.  
Substituting GPT-5 with GPT-4.5-Haiku preserved overall learning trends, indicating model-agnostic behaviour across same-scale LLMs.

\end{document}